\def\year{2022}\relax
\documentclass[letterpaper]{article} 
\usepackage{aaai22}  
\usepackage{times}  
\usepackage{helvet}  
\usepackage{courier}  
\usepackage[hyphens]{url}  
\usepackage{graphicx} 
\graphicspath{ {./imgs/} }
\usepackage{booktabs}
\usepackage{natbib}  
\usepackage{caption} 
\DeclareCaptionStyle{ruled}{labelfont=normalfont,labelsep=colon,strut=off} 
\usepackage{algorithm}
\usepackage{algorithmic}

\usepackage{newfloat}
\usepackage{listings}
\floatstyle{ruled}
\newfloat{listing}{tb}{lst}{}
\floatname{listing}{Listing}
\title{An Ensemble Approach to Acronym Extraction using Transformers}
\author{
    Prashant Sharma\equalcontrib\textsuperscript{\rm 1}, Hadeel Saadany\equalcontrib\textsuperscript{\rm 2}, Leonardo Zilio\textsuperscript{\rm 2}, \\
    Diptesh Kanojia\textsuperscript{\rm 2}, Constantin Orăsan\textsuperscript{\rm 2}\\
}
\affiliations{
    \textsuperscript{\rm 1}Hitachi CRL, Japan.\\
    \textsuperscript{\rm 2}Centre for Translation Studies, University of Surrey, United Kingdom.\\
    prashaantsharmaa@gmail.com, h.saadany@surrey.ac.uk\\ 
}

\begin{document}

\maketitle

\begin{abstract}
Acronyms are abbreviated units of a phrase constructed by using initial components of the phrase in a text. Automatic extraction of acronyms from a text can help various Natural Language Processing tasks like machine translation, information retrieval, and text summarisation. This paper discusses an ensemble approach for the task of Acronym Extraction, which utilises two different methods to extract acronyms and their corresponding long forms. The first method utilises a multilingual contextual language model and fine-tunes the model to perform the task. The second method relies on a convolutional neural network architecture to extract acronyms and append them to the output of the previous method. We also augment the official training dataset with additional training samples extracted from several open-access journals to help improve the task performance. Our dataset analysis also highlights the noise within the current task dataset. Our approach achieves the following macro-F1 scores on test data released with the task: Danish (0.74), English [Legal] (0.72), English [Scientific] (0.73), French (0.63), Persian (0.57), Spanish (0.65), Vietnamese (0.65). We release our code and models publicly\footnote{\url{https://github.com/dipteshkanojia/PR-AAAI22-SDU-ST1-AE}}.
\end{abstract}

\section{Introduction}
\label{sec:intro}

Acronyms are commonly used to shorten known units of text in various domains such as scientific~\citep{pustejovsky2001automatic}, medical~\citep{dannells-2006-automatic}, business~\citep{menard2011classifier} and legal~\citep{tsimpouris2015acronym}. Humans can usually identify acronyms in a text without too much difficulty by relying on various surface clues. Without knowing the meaning of acronyms, it is not possible to understand a text properly. Moreover, in absence of the long forms of an acronym translators and interpreters may have difficulties translate a text reliably.

Automatic identification of acronyms and their corresponding long forms is a relevant issue in the domain of Natural Language Processing (NLP) as it can help tasks such as information extraction and retrieval~\citep{sanchez2011automatic,ballesteros1996dictionary}, machine translation~\citep{kirchhoff2016unsupervised} and also machine interpreting~\citep{braun2019technology}. Acronyms represent relevant parts of the text and can confuse translation models. Thus, the task of automatically identifying and extracting acronyms from a text can be challenging for a machine.
\begin{figure}[t!]
    \centering
    \includegraphics[width=1\columnwidth]{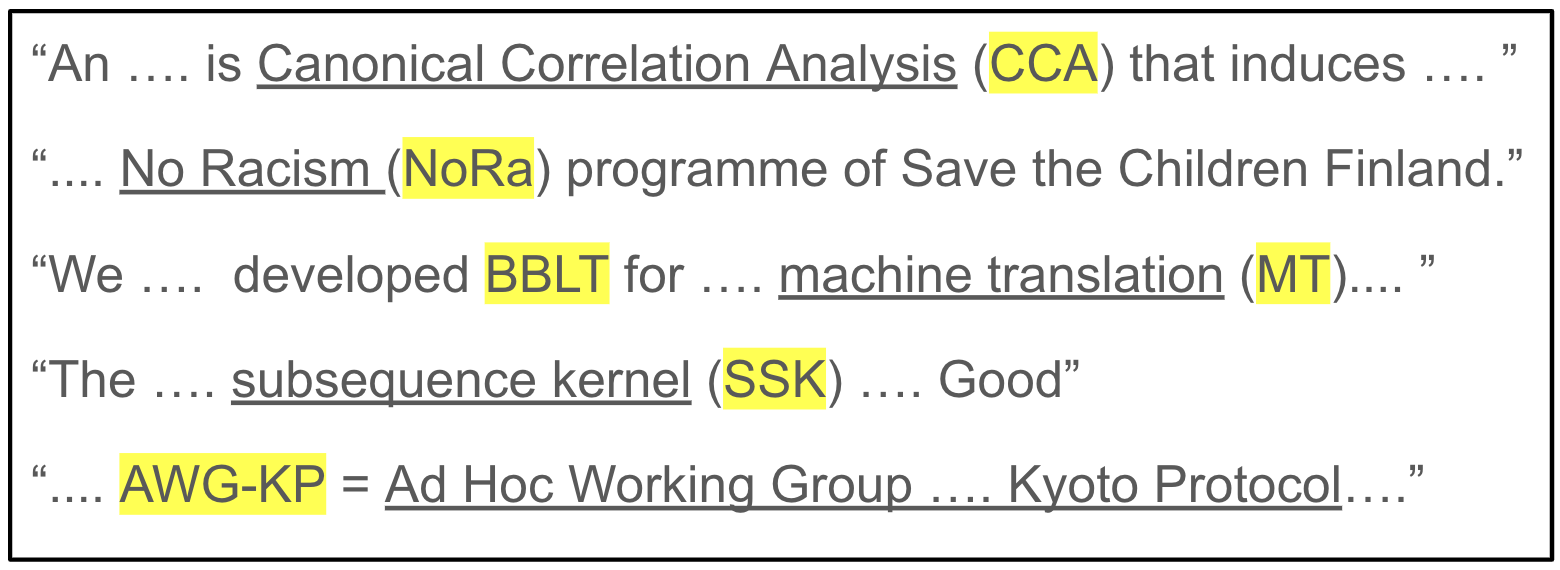}
    \caption{Examples of Long Forms and their Acronyms}
    \label{fig:example}
\end{figure}
The shared task-1 for acronym extraction (AE) under the Scientific Document Understanding (SDU) workshop\footnote{https://sites.google.com/view/sdu-aaai22/home} allows researchers to tackle this challenge and propose novel ways to solve it. The task requires participants to submit systems that can automatically identify acronyms and their long forms from a given piece of text. Figure~\ref{fig:example} shows two examples of acronyms and their long forms from the English dataset (Scientific domain) and English dataset (Legal domain). These examples are from the task dataset provided as-is to the participants. The acronyms in parentheses are highlighted in yellow, and their respective long forms are underlined. Automatic acronym extraction can be particularly challenging and based on our analysis of the training data for this task~\citep{veyseh-et-al-2022-Multilingual}, we observe: 
\begin{itemize}
    \item Acronyms are known to be present in uppercase letters in the text, but there can be instances where they can contain lowercase letters (\textit{e.g.}, CATiB, SEqui, OEqul, SRals ing)
    \item There can be instances where the long forms of the acronyms are not present in the text (\textit{e.g.,} We originally developed BBLT for ourselves..)
    \item There can be different long forms for the same acronym (\textit{e.g.,} CT has been used as an acronym for both ``contract'' and ``certain'')
    \item Acronyms can have special characters present in them (\textit{e.g.,} Communist Party of the Philippines-New Peoples Army-National Democratic Front (CPPNPA-NDF) )
    \item Multiple acronymised letters are part of the same word (\textit{e.g.,} subsequence kernel (SSK), maximum entropy (MaxEnt) )
\end{itemize}
Apart from the challenges discussed here, we also noted noise in the dataset, which we discuss in the dataset analysis subsection later (Section~\ref{subsec:dataAnalysis}). However, for the challenges discussed here, rule-based approaches fail, in particular, as they try to generalise over a pattern or a regular expression to detect acronyms from the text. There are multiple outliers that cannot be detected with the help of such approaches, as can be seen in the results of the rule-based approach implemented by the organisers of this task as a baseline. We discuss this approach in brief in Section~\ref{sec:approaches}. For this reason, we focused our efforts to develop a data-driven approach (more precisely a deep learning-based approach) to extract acronyms from the multilingual dataset prepared by the organisers of the shared task. 

In this paper, we describe our efforts to create a system that can extract acronyms and their long forms from a multilingual dataset. We model the task of acronym extraction as a sequence labelling problem and perform token classification considering each dataset sample as a sequence. After experimenting with several architectures, we decided to use an Ensemble approach which relies on two methods. The first method utilises the Transformer-architecture-based multilingual language model, XLM-R~\cite{conneau2019unsupervised}, to perform fine-tuning and extract acronyms. We also perform the acronym extraction task with the help of a Convolutional Neural Network (CNN), which employs word embeddings from a different source, as described later in our work. The resultant outputs from both these methods are then combined to create an ensemble output which we submit to obtain our scores for the task.

\section{Related Work}

The task of extracting acronyms from text has been performed in different domains for English, with most of the approaches being rule-based approaches ~\citep{taghva1999recognizing,yeates1999automatic,park2001hybrid,larkey2000acrophile}.~\citet{schwartz2002simple} implemented an algorithm for identifying acronyms by using parenthetical expressions as a marker of a short form. Their work is based on a previous work by~\citet{pustejovsky2001automatic} which also extracts acronyms using a similar method.~\citet{dannells-2006-automatic} performs the extraction of acronym-definition pairs from Swedish medical texts by primarily using a rule-based approach to extract acronyms and then a memory-based supervised machine learning approach to compare and evaluate the results. A rule-based approach was also implemented by~\citet{okazaki-ananiadou-2006-term} for term recognition, and it discusses the extraction of acronyms and their long forms. This system mines acronyms based on parenthetical expressions as a marker of a short form as previous methods had described. However, for mining long forms, they created a candidate list based on frequent co-occurrences of word sequences.~\citet{movshovitz2012alignment} investigate the use of Hidden Markov Model (HMM) for the extraction of acronyms from text.~\citet{ehrmann2013acronym} show how acronym recognition patterns, initially developed for medical terms, can be adapted to the more general news domain. Their efforts led to automatically merging the numerous long-form variants referring to the same short form while maintaining non-related long forms separately. Their work is based on the algorithm developed by~\citet{schwartz2002simple}, but they perform the task of acronym extraction for 22 languages. 

Machine learning-based approaches for the extraction of acronyms have been utilised in many previous studies~\cite{nadeau2005supervised,kuo2009bioadi}. With the advancement of research in NLP, various methods to extract word embeddings for text have been proposed, the most recent of them being contextual language models. To detect acronyms without local definitions,~\citet{rogers2021ai} applied two deep learning approaches: bi-directional LSTM with CRF and Transformer models.~\citet{li2021systems} utilise transformer-based architecture for modelling the task of acronym identification as a sentence-level sequence labelling problem.~\citet{zhu2021bert} incorporate the FGM adversarial training strategy for fine-tuning BERT for robust and generalised acronym identification. This was the winning system for the Acronym Extraction Shared Task held at SDU workshop in 2021.

In this paper, we employ the previously proposed approach of fine-tuning a language model for the task of acronym extraction~\citep{kubal2021effective}. However, we extract additional data from PLOS journals and perform additional data analysis. We describe data augmentation and pre-processing techniques in the upcoming sections.

\section{Dataset}
\label{sec:data}

The dataset provided by the task organisers consists of independent sentences in Danish, English, French, Persian, Spanish, and Vietnamese languages in the JSON format~\citep{veyseh-et-al-2022-MACRONYM}. The English dataset is further divided into two different domains: legal and scientific. The dataset statistics can be seen in Table~\ref{tab:datastats}. However, each text sample can contain multiple lines of the text, thus containing up to 1050 words in each sample as observed from the training dataset. 

\begin{table}[t!]
\centering
\resizebox{\columnwidth}{!}{%
\begin{tabular}{@{}llll@{}}
\toprule
 & Training & Development & Test \\ \midrule
Danish & 3082 & 385 & 386 \\
English (Legal) & 3564 & 445 & 446 \\
English (Scientific) & 3980  & 497 & 498 \\
French & 7783 & 973 & 973 \\
Persian & 1336 & 167  & 168 \\
Spanish & 5928 & 741 & 741 \\
Vietnamese & 1274  & 159 & 160 \\ \bottomrule
\end{tabular}%
}
\caption{Dataset Statistics, in terms of number of dataset samples, for the Acronym Extraction task as provided by the task organizers.}
\label{tab:datastats}
\end{table}
\begin{figure*}[ht!]
    \centering
    \includegraphics[width=1\textwidth]{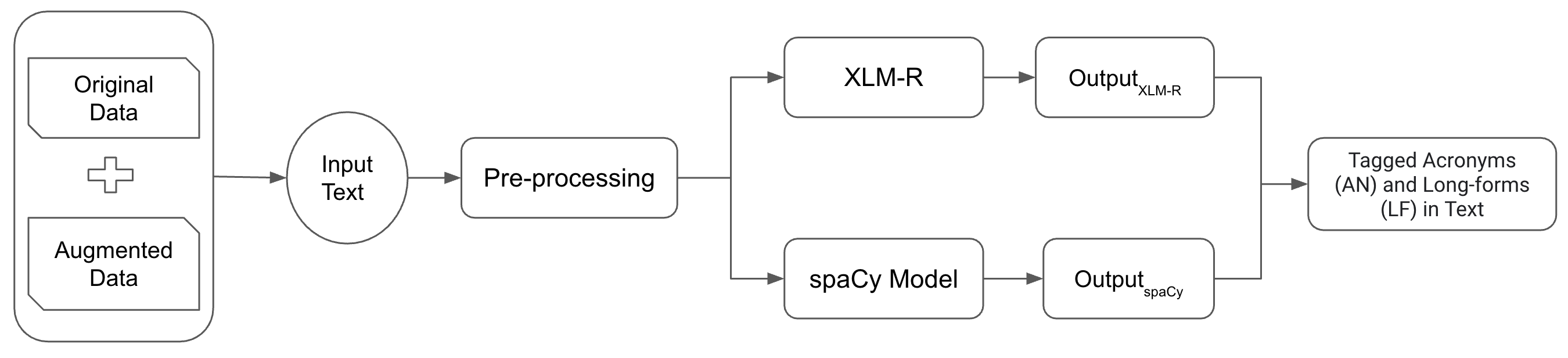}
    \caption{System Architecture for Our Ensemble Approach}
    \label{fig:sysarch}
\end{figure*}

\subsection{Dataset Preprocessing}
\label{subsec:preprop}
The fact that we model the task as a sequence labelling problem required us to convert the text provided into a BIO (short for \textbf{B}eginning, \textbf{I}nside, \textbf{O}utside) format. BIO is a common format for tagging a token in a chunking or a named entity recognition task in computational linguistics~\citep{Ramshaw1999}. We convert each JSON task dataset into the BIO format with the help of a custom script written in Python. The custom tags we use to convert each token in the sequence resemble Named Entity tags and result in the following list of tags: [\textbf{O} (Outside), \textbf{B-AN} (Begin Acronym), \textbf{I-AN} (Inside Acronym), \textbf{B-LF} (Begin Long Form), \textbf{I-LF} (Inside Long Form)].

\subsection{Dataset Analysis}
\label{subsec:dataAnalysis}
We analysed the dataset provided for the task and observed the following issues:
\begin{itemize}
    \item There are instances with missing or incomplete annotation (e.g. the acronym SDI has associated with the long-form ``selective dissemination'' instead of ``selective dissemination of information'', which is present in the instance).
    \item Segments with wrong annotation (e.g. ``US\$ 3'' was annotated as an acronym of ``USh 3,000'', when these are actually conversions between two currencies).
    \item Segments with over-annotation (e.g. the acronym ``(GHS)'' had the brackets included in the annotation).
\end{itemize}

In an analysis performed on the first 100 instances of the English scientific training dataset, we found 28 instances with such issues. These issues were also found in 21 instances among the first 100 instances of the English legal training dataset. The presence of such a high number of errors in the dataset poses some serious challenges to any data-driven method.

\subsection{Data Augmentation}
\label{subsec:dataAugm}
To increase the amount of training data, a separate training dataset was created by using information extracted from the PLOS\footnote{\url{https://plos.org/}.} open-access journal publications. The XML versions of these publications are freely distributed along with the PMC Open Access Subset \footnote{\url{https://www.ncbi.nlm.nih.gov/pmc/tools/openftlist/}.} and amount to 305,445 texts.

The XML files in the PLOS corpus have a section for abbreviations, where all abbreviations used in a paper are made explicit along with their long forms. This information was used to extract sentences from research articles within the PLOS corpus. For the augmented dataset, only sentences containing at least one abbreviation with a related long-form were included (other abbreviations with or without long forms could be present as well). This augmented training dataset contained a total of approx. 93k samples for English (Scientific domain). 

\section{Our Approach}
\label{sec:approaches}

The baseline approach provided the task organisers uses a single rule, i.e., if the word inside a parenthesis contains more than 60\% uppercase letters, it is to be identified as an acronym. Moreover, the number of uppercase letters in this acronym form a sliding window for words before/after the acronym. If each uppercase letter matches the first characters of words in the sliding window, the words constitute the long-form phrase. This approach clearly fails to address the challenges discussed in Section~\ref{sec:intro} of the paper. The resulting macro-F1 scores for the \textbf{baseline approach on the development} data are: English (Legal) - 0.1258, English (Scientific) - 0.1084, Danish - 0.0950, French - 0.0806, Spanish - 0.0831, Persian - 0.4437, Vietnamese - 0.3538. Hence, we needed to look for alternative methods for identifying acronyms and their long forms. After considering a number of options, we decided to experiment with a deep-learning-based ensemble approach. 

\textbf{Our Ensemble Approach:} We concatenate all the multilingual task data into a single training dataset. This was then concatenated with the PLOS data described above to increase the train data size to approx. 113k dataset samples. This data is used to fine-tune the multilingual language model described below to perform the acronym extraction (AE) task. We obtain the test output from two different methods described below and concatenate them. The fine-tuning method described below is able to label both acronyms and long forms. The spaCy blank model-based method described below is only able to obtain acronyms for the task but helps achieve improved F1 scores for the task. The architecture for our approach is shown in Figure~\ref{fig:sysarch}.

\subsection{Fine-tuning with XLM-RoBERTa}
XLM-RoBERTa (XLM-R)~\citep{conneau2019unsupervised} is a multilingual contextualised Language Model (LM) pre-trained on filtered CommonCrawl data from 100+ languages. Each language dataset from this task is included in this model\footnote{\url{https://github.com/facebookresearch/XLM}}. 

Our approach utilises this transformer architecture-based pre-trained LM and fine-tunes it for the downstream sequence labelling task. The LM used for our approach is XLM-R$_{base}$, and has approximately 270M parameters with 12-layers, 768 hidden states, 3072 feed-forward hidden states, 8 heads; and is pre-trained on CommonCrawl data in over 100 languages. The fine-tuning process adds a hidden linear layer on top of the pre-trained LM and projects the output to a \textit{softmax} layer for token classification. We perform further hyperparameter tuning as described below.

We observed that during the inference phase, the output token length was truncated to 128 tokens as it was the default `maximum sequence length' parameter. To preserve the entire token length, the two parameters, namely, `sliding window' and `maximum sequence length' were being used with the original model. The sliding window prevents the truncation of sentences by splitting the input sequence into multiple windows if it exceeds the default maximum sequence value. The sliding window problem represents the broken contextual information while predicting token class, and hence it was not used. We carried out multiple experiments with the `maximum sequence length' parameter and observed that the model performed the best when it was limited to 512. The other values for maximum sequence length we experimented with were 128, 256, 350, 450, and 512. The language model was fine-tuned by using the fairSeq~\citep{ott2019fairseq} library. We used an NVIDIA Quadro RTX 5000 GPU with 16 GB of memory for carrying out these experiments. This approach utilizes the training data to learn how to classify each token as B-AN, I-AN, B-LF and I-LF, in the text. With the help of further post-processing using a custom script in Python, we were able to convert the predictions in the JSON format as required for the evaluation phase of the task. 

\subsection{spaCy Blank Model}
\label{subsec:spacy}

The other model that we used for AE was a spaCy v3.2\footnote{\url{https://spacy.io/models}} blank NER model. The spaCy v3.2 model is based on predicting context-sensitive vectors for each word in the input by a token-to-vector model. The embeddings in this model are obtained from the Bloom embeddings where each sub-word is transferred into a string of fixed symbols (e.g. 0-9 integer transferred to letter d, capital letter to W and lower case letter to w)~\citep{blooming}. This strategy has proven to be effective in handling out-of-vocabulary (OOV) tokens; instead of dumping all OOVs in one bucket, each OOV is given a unique representation. The method offers a variety of neural architectures for building a blank NER model to predict task-tailored entities. We opted for the trigram-CNN architecture learning via a transition-based approach which takes a window of the embeddings on either side of each word in the sentence and concatenates them in a multi-layered perceptron followed by an attention layer~\citep{lample2016neural}. A Maxout Unit~\citep{goodfellow2013maxout} is used as an activation function that calculates the `maximum' of the inputs. These architecture parameters have performed well for NER tasks\footnote{\url{https://v2.spacy.io/usage/facts-figures}}.

For training, we use the English (Scientific) and English (Legal) training sets consisting of 7523 instances and evaluate our method on the English (Scientific) and English (Legal) development sets. We use a limited training dataset as these experiments are performed with CPU cores. Before training, we pre-processed the data to conform with the `.spacy' format, where each positive instance was assigned the NER label acronym along with its specific indices. For the spaCy pipeline parameters, we chose the spaCy en\_core\_web\_sm model, which is the small model trained on written English language web text (blogs, news, comments) including vocabulary, vectors, syntax and entities. We trained this model on a CPU with 100 iterations and with a batch size of 1000. 

We used this model as a zero-shot model for the AE task with the test sets of the other four languages during the final stage of the shared task. A note on the reason for choosing this blank spaCy model for AE in this task is that it has a CPU-optimised pipeline, and it is much cheaper to run than pre-trained models. Due to its competitive results to the more expensive pre-trained models, we plan to explore training a spaCy model with more data for future AE task.

\section{Results and Discussion}
\label{results}

\begin{table}[t!]
\centering
\resizebox{\columnwidth}{!}{%
\begin{tabular}{llll}
\hline
 & F1 & P & R \\ \hline
Danish & 0.74 & 0.78 & 0.70 \\
English (Legal) & 0.72 & 0.75 & 0.69 \\
English (Scientific) & 0.73 & 0.77 & 0.69 \\
French & 0.63 & 0.68 & 0.59 \\
Persian & 0.57 & 0.64 & 0.51 \\
Spanish & 0.65 & 0.65 & 0.65 \\
Vietnamese & 0.65 & 0.64 & 0.66 \\ \hline
\end{tabular}%
}
\caption{Results obtained using our ensemble approach over the test data as provided for the task where P is Precision, R is Recall, and F1 is the Macro-F1 score as used for the task.}
\label{tab:results}
\end{table}

Using the fine-tuned XLM-R model, we obtained acronyms and long forms for the test data provided for this task. We then concatenated this output with the output from the spaCy blank model. The results obtained for the final test set output are present in Table~\ref{tab:results}. Our approach was outranked by several other systems submitted for the task, but we show a significantly improved set of results over the baseline method proposed for the task. Our results also show how the AE task can be modelled as a sequence labelling problem, thus utilising pre-existing architecture for the NER problem in NLP. The performance of our approach in comparison to other systems submitted at the task was comparatively lower, which can be attributed to the fact that we use a single multilingual training model for all the languages. We also use a rather simple fine-tuning based approach and do not add a more sophisticated neural networks-based architecture. 

The output obtained using the fine-tuning method described above outperforms the rule-based baseline approach by a significant margin. This method helps our approach gain significant percentage points for both acronyms and long forms as compared to the baseline approach. However, we observed that this method did not recognise many acronyms, resulting in low recall values. We also observed that due to the noise present in the data, this method tagged special characters like `)' as a part of the acronym. We also observed that the fine-tuning process tagged a lot of \textit{stopwords} as long forms even when they were not a part of any long-form sequence (\textit{e.g.}, for, the, of). We post-process the output of this model to rectify such errors.

The performance of the spaCy method, however, is significantly better at extracting acronyms as it uses a CNN architecture and performs well for the English language. However, it needs to be pointed out that, despite the fact that the dataset was multilingual in nature, most instances of long forms and acronyms were present in the English language. The results of the spaCy model for the extraction of only acronyms against the development set were: English Scientific - Precision: 0.8847, Recall: 0.7990, F1: 0.8397; and English Legal - Precision: 0.9168, Recall: 0.7354 and  F1: 0.8161. As noted, this method only extracts acronyms and does not work well with long forms.

In most of the cases, we observed that our approach obtains higher precision than recall. This observation is expected as the fine-tuning process expects a lot more `O' tokens compared to the AN or the LF classes. Our approach, which classifies each token, confuses a lot of `AN's as `O's. In fact, when the output of the spaCy model was deconcatenated, we observed that the XLM-R based method for the Persian language had only achieved an F1 of 0.27 compared to the overall F1 of 0.57. The task performance of our approach on the Spanish test data, however, is an exception as it shows a steady Precision, Recall, and F1 scores of 0.65.

\section{Conclusions and Future Work}
\label{sec:conc}

In this paper, we propose a deep-learning-based approach to extract acronyms and long forms from the data provided for the task. We discuss the problem of acronym extraction and show how challenging it is to accomplish this task automatically. The dataset provided for the task is multilingual in nature, and our approach attempts to build a single model which can handle all the languages. However, our dataset analysis shows that the data provided for the task should indeed be manually validated first to remove the noise. We also augmented this dataset with more samples from the PLOS open-access journal to improve the dataset size, but it is unclear how much this helped improve the performance. We modelled this AE task as a sequence labelling problem and used an ensemble approach which utilises two different methods: (1) based on fine-tuning the XLM-R model to extract ANs and LFs, and (2) based on using a CNN architecture provided by spaCy blank modelling method. Our results significantly outperform the baseline results and show that this approach does work for the AE task. We release the code and the models used for this task\footnote{\url{https://github.com/dipteshkanojia/PR-AAAI22-SDU-ST1-AE}}.

For the future, we aim to perform this multilingual AE task by separating the models individually for each language. We plan to use the data from PLOS to augment each training dataset and perform further experiments. We also plan to collect more data for each language for the task and augment it with the training data for each of the models. With this augmented resource, we plan to perform an extensive analysis of the acronym extraction task and present our findings in the near future. Our eventual goal is to perform exhaustive experimentation with various datasets/methods and empirically find the best performing approach for this task. 

\bibliography{aaai22}

\end{document}